\documentclass[runningheads]{llncs}
\usepackage{graphicx}
\usepackage{float}
\usepackage{amsmath}
\usepackage[hidelinks]{hyperref}
\usepackage{multirow}
\usepackage[table,xcdraw]{xcolor}
\usepackage{amssymb}

\begin{document}
\title{A LayoutLMv3-Based Model for Enhanced Relation Extraction in Visually-Rich Documents}
\titlerunning{A LayoutLMv3-Based Model for Relation Extraction}

\author{Wiam Adnan\inst{1,2} \and
Joel Tang\inst{2} \and
Yassine Bel Khayat Zouggari\inst{2}\and \\
Seif Edinne	Laatiri\inst{2} \and
Laurent Lam\inst{2} \and
Fabien Caspani\inst{2}}
\authorrunning{W. Adnan et al.}
%
\institute{Ecole Polytechnique, France \and
BNP Paribas CIB, France\\
\email{wiam.adnan@polytechnique.edu}\\
\email{\{firstname.lastname\}@bnpparibas.com}}

\maketitle              

\begin{abstract}
Document Understanding is an evolving field in Natural Language Processing (NLP). In particular, visual and spatial features are essential in addition to the raw text itself and hence, several multimodal models were developed in the field of Visual Document Understanding (VDU). However, while research is mainly focused on Key Information Extraction (KIE), Relation Extraction (RE) between identified entities is still under-studied. For instance, RE is crucial to regroup entities or obtain a comprehensive hierarchy of data in a document. In this paper, we present a model that, initialized from LayoutLMv3, can match or outperform the current state-of-the-art results in RE applied to Visually-Rich Documents (VRD) on FUNSD and CORD datasets, without any specific pre-training and with fewer parameters. We also report an extensive ablation study performed on FUNSD, highlighting the great impact of certain features and modelization choices on the performances.

\end{abstract}

\section{Introduction}

In today's academic and industrial world, the sheer volume of unstructured textual information poses a challenge to knowledge extraction and comprehension from documents. The ability to understand complex relationships between entities within vast documents is important in extracting valuable insights and structuring information. This is applicable to documents in diverse domains, ranging from biomedical research and financial analysis to legal documentation and news articles. Therefore, there is a need to predict attributes and relations for entities in a sequence of text. This is the definition of the Relation Extraction (RE) task, emerging as instrumental in the process of understanding complex unstructured documents.\\

RE task is usually preceded by an Entity Extraction (EE) step, which aims at identifying spans of typed information in unstructured text. While recent years have witnessed outstanding breakthroughs in the field of EE, particularly through Transformer-based pre-training methods, RE, on the other hand, received less attention. Noticeably, very few works focus on Visually-Rich Documents (VRD), and often propose a modified pre-trained scheme to adapt model embeddings for RE task \cite{hong2022bros,geolayoutlm}.\\

In this study, we explore a variety of approaches to leverage geometric and semantic information to improve RE modeling with pre-trained Transformers, in particular LayoutLMv3 \cite{layoutlmv3} backbone, without additional pre-training. This approach offers significant flexibility, as the absence of further pre-training allows for easy adaptation to other existing multi-modal backbones, thus expanding its applicability in understanding VRD. Additionally, we conduct an extensive ablation study to investigate the impact of multiple elements on the performance of the RE model. These elements encompass various aspects, including document's block ordering, model properties, and multi-task learning, providing valuable insights into the contributions of these factors and illuminating potential avenues for future research.\\

We summarize our contributions as follows:

\begin{enumerate}
    \itemsep 1em
    \item We introduce a methodology that achieves performance levels equal to or better than the current state-of-the-art in RE tasks for VRDs, all without the need for specific geometric pre-training and with a reduced number of parameters.
    \item We present an extensive ablation study that illustrates the effects of different training setups and additional features specific to VRDs, offering a comprehensive understanding of the factors influencing the RE model's performance.
\end{enumerate}

\section{Related works}

The objective of RE is to predict relations between entities in a text \cite{huang-wang-2017-deep}. Such relations can be typed or not, and are usually directed (asymmetrical relationships). This task is to distinguish from Entity Linking (EL), where the objective is to find a relation between an instance appearing in the text with an entity canonically appearing in a knowledge base \cite{fanghiqhquality,fangjointentity}. Noticeably, in Natural Language Processing (NLP), RE is explored in two significantly different setups, either on purely textual inputs or VRDs. While our work focuses on the latter, we present key related works of both fields.\\

The field of RE has traditionally focused on the textual modality, evolving from sentence-level to document-level complexities. At sentence-level, the task is usually to identify and classify relationships between one pair of entities \cite{zhang2017tacred}. The modelization usually relies on pre-trained language models to obtain entity representations and apply a classifier on them. If most of the existing work propose new pre-training procedures to enhance the downstream RE task \cite{wang-etal-2022-deepstruct,ma2023dreeam}, others propose introducing additional guiding features to label entity span and type. This approach improves the representation of entities by encoding their type, thus improving the task of RE \cite{zhou-chen-2022-improved}. Extending the possibility of relationships between entities, document-level entity RE exhibits specific challenges. Document-level RE is applied on usually longer documents containing several pairs of linked entities. Also, one named entity can appear multiple times in a document and be related to different other entities, which is an additional difficulty for the model, as it requires an understanding of the context of each entity occurrence \cite{zhou2021atlop}.\\

Text-based RE methods mainly rely on language models as a backbone originating from BERT \cite{devlin2019bert}, which are inherently tailored for textual data. However, in VRDs, there is a need for models able to encode a variety of modalities. Emerging language models, such as LayoutLM \cite{Xu_2020}, LayoutLMv2 \cite{xu2022layoutlmv2}, LayoutLMv3 \cite{layoutlmv3} and BROS \cite{hong2022bros}, address this limitation by taking into account layout (bounding boxes), position (positional encoding or embedding for each token), and image information in addition to text, thus providing a versatile framework for relationship extraction in VRDs.\\

The geometric information included in the document layout represents important features to be considered for the ER task in VRD. These features encompass not only the 2D \textit{absolute} spatial positions of text elements on a document page, but also imply \textit{relative} spatial positions, which describe how text segments are positioned with respect to each other. Models designed for VRD typically incorporate the geometric information as input, and use various approaches to capture and learn both \textit{absolute} and \textit{relative} spatial features. \\

Models in the LayoutLM series \cite{layoutlmv3,xu2022layoutlmv2,Xu_2020}, focus only on accurately encoding the 2D \textit{absolute} positions of tokens using pre-training tasks that align text with layout \cite{layoutlmv3}. To induce the \textit{relative} positions of tokens, these models rely on 1D sequential reading order, often determined by OCR output. In contrast, other approaches emphasize pre-training to comprehend directly both \textit{absolute} and \textit{relative} spatial features of textual segments using geometric pre-training objectives \cite{li2021structext,geolayoutlm}. Such approaches afford a deeper understanding of spatial relationships, potentially improving performance on VRD tasks. Additionally, methods like BROS \cite{hong2022bros} adapt the attention mechanism to directly leverage the \textit{relative} positions of bounding boxes, offering another avenue for encoding spatial relationships effectively. \\

GeoLayoutLM, which is based on a BROS backbone architecture, is currently the state-of-the-art model for RE task \cite{geolayoutlm}. The authors proposed a combination of objective pre-training features, aiming to teach the model inductive biases regarding document geometry, focusing on vertical and horizontal segment alignments. They also propose to jointly optimize the semantic EE and RE tasks, since knowledge of entity types is an important feature for RE.\\

Applications of RE methods on VRDs are still under-studied. In particular, there is little work studying meaningful features and their impact on metrics. Moreover, recent work requires pre-training which is computationally expensive, while being not a strict requirement. Hence, in our work, we focus on methods that don't require further pre-training and are widely available.

\section{Approach}

In this section, we describe and formalize the tasks of EE and RE within VRDs. Subsequently, we detail the architecture employed for RE. We then explore various methods aimed at enhancing the semantic understanding of entities and strategies to improve comprehension of their spatial relationships.

\subsection{Task formalization}

In this subsection, we present \textbf{matrix prediction}, formalizing the task of predicting relationships in entities belonging to a sequence. \textbf{Sequence} (of tokens) denotes text and any aligned multi-modal input, such as layout information attached to tokens. We also suppose having image features that are fed to the model at the same time without being explicitly aligned with the sequence. An \textbf{entity} is defined by a contiguous span of tokens. We admit having an oracle defining such spans of tokens so that we have a delimitation of the entities on which we predict their relationships with others. Then, for each pair of entities, the model is trained to predict if there is a relation between the first and the former. This method aims to predict the relations between all pairs of entities in the document in a single pass. In the literature, similar matrix-based approaches have been employed by models such as StrucText \cite{li2021structext} and GeolayoutLM \cite{geolayoutlm}.\\

In the matrix-based approach, relations between entities are represented using a $n \times n$ matrix denoted $M$. Each entry $M_{i,j}$ is binary, indicating the relation between entities $i$ and $j$. For key-value pairs, $M_{i,j} = 1$ indicates entity $i$ is the key to entity $j$. This matrix is often asymmetrical due to the directional nature of such relationships. Conversely, when considering groups, $M_{i,j} = 1$ if entities $i$ and $j$ belong to the same group, leading to a symmetric matrix, as $M_{i,j} = M_{j,i} = 1$ for all $i$ and $j$.\\

Considering entities as nodes in a graph, the relation matrix $M$ serves as an adjacency matrix, representing various relationship types, including hierarchical structures like trees or undirected groups. This approach eliminates the need to select a specific label to represent a group, a distinction made for problem simplification in the first approach.

\subsection{Architecture} 
\label{architecture}

\begin{figure}[t]
\centering
\includegraphics[width=\textwidth]{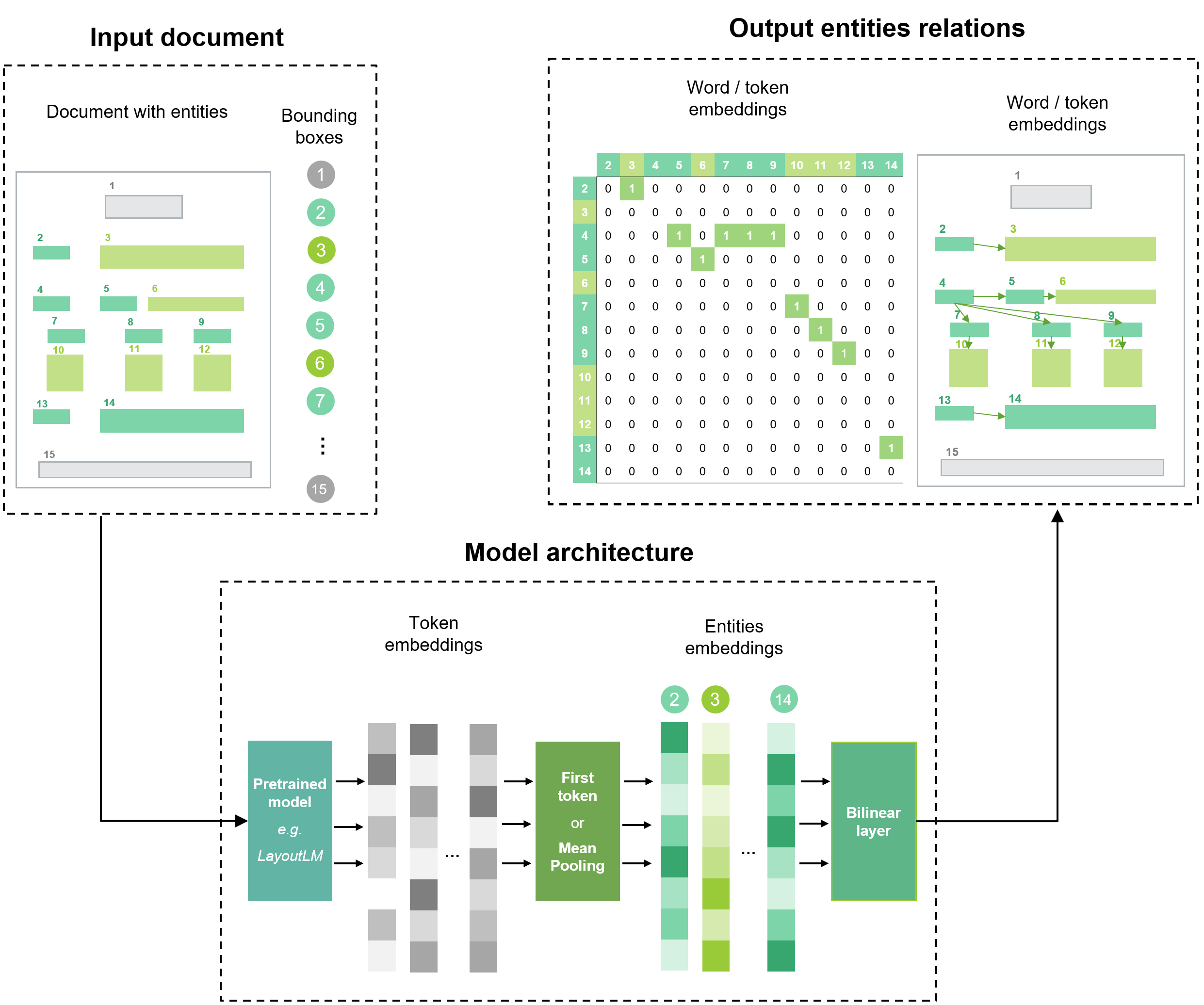}
\label{fig:architecture}
\caption{Illustration of the modelized Relation Extraction (RE) task from a Visually-Rich Documents (VRD). Text is split into bounding boxes and follows an order given by the OCR. The model is end-to-end trained to predict a relationship matrix via an asymmetric bilinear layer.}
\end{figure}

We choose to use LayoutLMv3 \cite{layoutlmv3} as our backbone model, which is a Transformer encoder-only architecture, taking text, spatial and visual modalities as inputs. We initialize our weights from the pre-trained LayoutLMv3 model, which was trained with a combination of three tasks: masked language modeling (MLM), masked image modeling (MIM) and word-patch alignment (WPA). We do not continue pre-training and will fine-tune directly from these prior weights. Additionally, the model will be fine-tuned on EE task as outlined in §\ref{subsubsec:E}, necessitating a definition of the architecture for both tasks.\\

The relation prediction head architecture begins by calculating entity embeddings, either using the first token embedding of an entity or the mean of all entity token embeddings. We then employ a bilinear layer for fusing entity representations post-linear transformation \cite{li2021structext} \cite{wang-etal-2020-docstruct}. Formally, assuming $E_i$ and $E_j$ to be the embeddings of entities $i$ and $j$ (one embedding per row), respectively and an asymmetric parameter matrix $M$ is used to extract the relationships between entities in a probabilistic form. The relationship from entity $i$ to entity $j$ is given by:
    \[ P_{i \rightarrow j} = E_j M E_i^T \] \label{eq:bilinear}\\
    where $P_{i \rightarrow j}$ is the probability that entity $i$ links to entity $j$. The final relation output, \( R \), is defined based on the relationship probability \( P_{i \rightarrow j} \) between entities \( i \) and \( j \). Specifically, the relation \( R_{ij} \) is given by:

$$ R_{ij} = \begin{cases} 
1 & \text{if } P_{i \rightarrow j} > 0.5 \\
0 & \text{otherwise}
\end{cases} $$\\

The relation prediction head is trained via the binary cross-entropy objective function.\\

EE head is a linear classification, taking as an input the first token of a span of tokens, followed by a softmax operation predicting probabilities of the span of tokens to belong to some class with IOB tagging \cite{brill-1992-simple}. During the fine-tuning phase for both tasks, we either optimize jointly by summing loss functions or focus solely on optimizing the RE task.

\subsection{Including Entity Types}
Incorporating entity type information is crucial for informing the model about potential relationships between entities. In this work, we assess two orthogonal techniques allowing the RE model to incorporate this information. First, it is possible to jointly optimize the EE task with the RE task by sharing the encoder weight and applying two distinct heads, one for each task. This method aims at reaching a shared representation for both tasks and allows to distil entity type information during the training \cite{geolayoutlm}. Second, inspired from work in the text-only setups, we test the impact of directly prepend entity types in natural language to entity spans \cite{zhou-chen-2022-improved}. 

\subsubsection{Joint EE and RE fine-tuning (EEF):}
\label{subsubsec:E}
It has been shown that jointly fine-tuning on EE and RE tasks can improve performances of the model on the RE task \cite{geolayoutlm}.
This dual-task training enriches the token embeddings. The embeddings generated by our model serve two purposes: identifying entities and their types, and predicting relationships between entities.\\

The EE head is composed of a classification layer (linear layer and softmax operation) applied on each token that needs to be classified. The classes follow IOB tagging convention \cite{ramshaw1995text}: non-entity tokens are classified as \texttt{other}, while each class is separately classified into two classes: \texttt{B-<CLASS>} and \texttt{I-<CLASS>}. The former indicates the first token of a sequence of tokens for a given entity and the latter indicates tokens following a beginning token. This tagging allows to differentiate two same-class entities following one after the other. The RE task head is defined as in Subsection \ref{architecture}.\\

Both tasks are trained with cross-entropy loss functions.\
This approach is simple, at the expense of additional parameters and computation if explicit EE is not required.

\subsubsection{Entity Marker (EM):} 
\label{subsubsec:T}
We integrate the types of entities within the structured input to guide the model in understanding potential entity relationships. This approach involves presenting entity information within the OCR output text, allowing the model to effectively process it. Previous studies \cite{zhou-chen-2022-improved,zhang-etal-2017-position,baldini-soares-etal-2019-matching,wang-etal-2021-k} explored these techniques in sentence-level RE, focusing primarily on scenarios involving a single referent or parent entity.\\ 

Related work introduced \textit{Typed Entity Marker (punct)} technique, which has shown promising results \cite{zhou-chen-2022-improved}. This method doesn't introduce any new special tokens and only uses punctuation to differentiate two different entities' roles in the relationship and delimit their start and end. Notably, it uses the actual word (in-vocabulary) denoting the entity type, thus preserving semantic understanding. This approach significantly outperforms methods based on introducing new  (out-of-vocabulary) tokens to indicate classes.\\

As an example, the formatted input sentence with the typed entity marker would be:\\

\texttt{@ * person * Bill @ was born in \# \^{} city \^{} Seattle \#}.\\

However, while related work require punctuation to indicate the asymmetric relationship between two entities (punctuation will vary, with \texttt{*} and \texttt{\^} respectively denoting the two ends of the relationship between two entities), our method, using a bilinear layer for classification as highlighted in \eqref{eq:bilinear}, allows us to simplify the entity marker as: \texttt{entity} \(\rightarrow\) \texttt{\textit{entity type} entity}. By analogy, the sequence will become:\\

\texttt{\textit{person} Bill was born in \textit{city} Seattle}.\\

A strong advantage of this method is that it effectively reduces the number of used tokens per prediction. With punctuation, adding the typed entity marker required at least two windows to input one document for 53\% of the samples in CORD, compared to 0\% with our proposed method.

Layout information of the appended tokens are set to be identical to the bounding box of the first token of the entity.

\subsection{Geometry and Spatial Understanding}

The LayoutLMv3 model was pre-trained on three distinct tasks: Masked Language Modeling, Masked Image Modeling, and Word-Patch Alignment. It is a standard pre-training approach for multi-modal models, where masked objective functions are applied on each modality and the former one is designed to align both modalities representations. However, it was not pre-trained on geometry-specific tasks like the GeoLayoutLM model \cite{geolayoutlm,li2021structext}. The LayoutLMv3 model operates sequentially and incorporates positional encoding. Consequently, the model relies more on positional encoding than layout encoding. The order in which entities are presented significantly influences the model's predictions.\\

A significant challenge in real-world scenarios stems from imperfect order information of text blocks. The absence of a defined reading order in OCR outputs often compromises the accuracy of positional encoding \cite{hong2022bros}. Furthermore, it is important to note that many documents feature relationships between spatially adjacent entities. Understanding the layout is imperative for interpreting these relationships, as it is much less sensitive to errors and contains enough information with text to infer reading order. However, the LayoutLMv3 backbone, not being directly pre-trained on geometry-specific tasks, does not fully harness the potential of geometric relationship information. In this subsection, we present several methods aiming at giving more importance to the layout versus explicit token ordering or mitigating the impact of ordering errors.

\subsubsection{Layout Concatenation (LC): }
\label{subsubsec:L} Given that LayoutLMv3 was not pre-trained on geometric tasks, layout information tends to be diluted in embeddings. We propose to create a skip-connection between token-level layout information and the RE head, by concatenating normalized absolute coordinates of each token with its associated hidden state before performing relation prediction using RE prediction head.

\subsubsection{Bounding Boxes Ordering (BBO):}
\label{subsubsec:O}

The order in which entities are presented significantly influences the model's predictions. The arrangement of entities in the model's input, especially when aligned with the natural reading order, is crucial for understanding potential relationships. Therefore, ordering boxes by their vertical positions can enhance the model's accuracy in generating predictions. This approach mimics the human method of processing information. This approach helps in dealing with limitations in OCR outputs where the reading order is not always clear, aiding the model in making more informed and contextually accurate predictions.

\subsubsection{Bounding Boxes Shuffling (BBS):} 
\label{subsubsec:S}
As previously mentioned, our model predominantly utilizes positional encoding over layout encoding. The layout, or coordinates, of bounding boxes contain crucial geometric information about the relationships between entities. To address this, we randomly permute the bounding boxes in each batch during fine-tuning. This process introduces artificial noise into the positional encoding of tokens as taken by the LayoutLMv3 backbone. Consequently, it forces the model to consider more spatial coordinates. This approach effectively reduces the model’s dependence on the order of entities and encourages a focus on their actual coordinates to deduce relationships. However, it is important to note that while we shuffle the bounding boxes, we maintain the order of tokens within each individual box or segment.

\section{Datasets}

\begin{figure}[t]
\centering
\includegraphics[width=\textwidth]{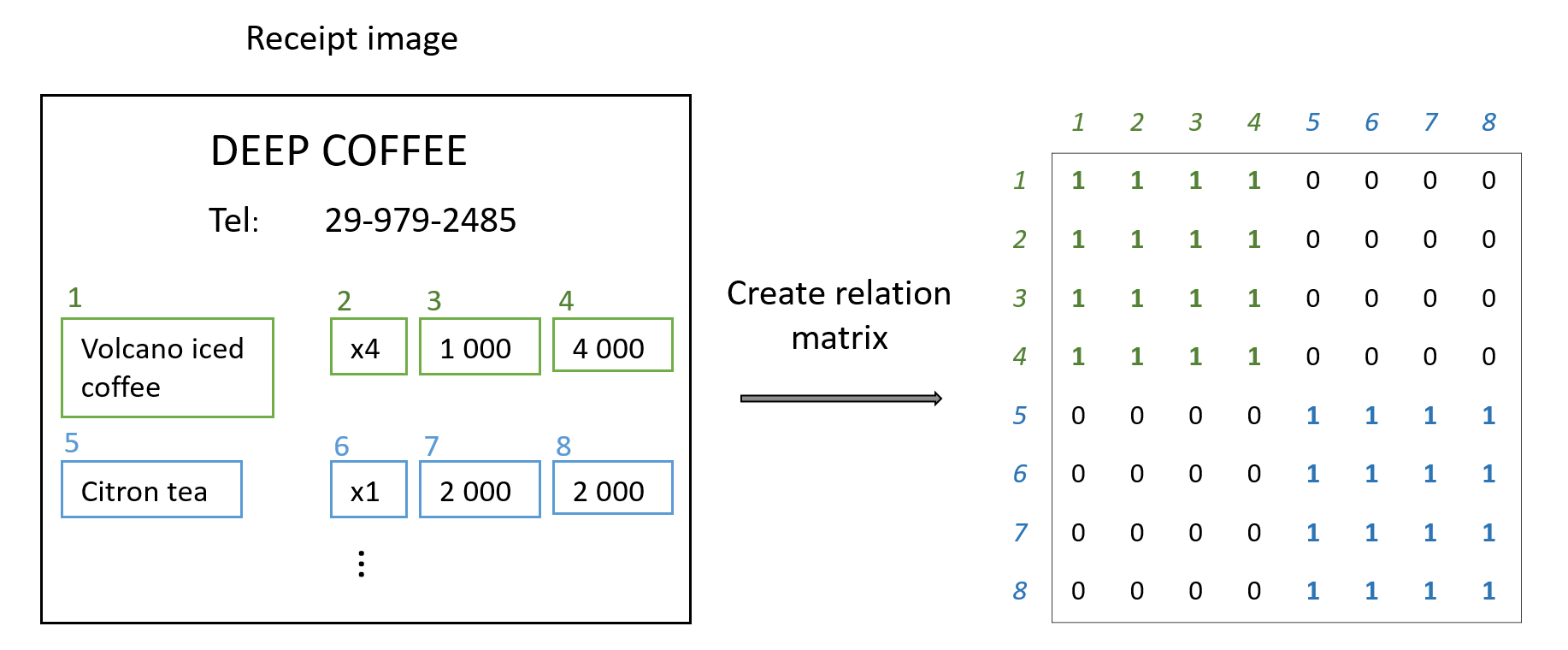}
\caption{Example of relations modeling in CORD dataset}
\label{fig:cord_re}
\end{figure}

\textbf{CORD (Consolidated Receipt Dataset)} \cite{park2019cord} is a dataset of restaurant receipts, which includes both images and box/text annotations for OCR, as well as multi-level semantic labels for parsing. Labels are structured into eight superclasses, which are further divided into 54 subclasses. The superclasses include store information, payment information, menu, void menu, subtotal, total, void total, and others. The CORD dataset consists of 1,000 annotated samples divided into training/validation/testing sets with the ratio 800/100/100. It is particularly suited to multimodal models that combine visual and textual information, as well as to the relation extraction task thanks to the multi-level semantic labels provided. Figure \ref{fig:cord_re} illustrates an example of a matrix with relevant groups in the CORD dataset.
\\

\textbf{FUNSD (Form Understanding in Noisy Scanned Documents)} \cite{jaume2019funsd} \cite{vu2020revisingfunsd} is a dataset designed for the extraction and structuring of textual content from forms. The forms originate from various domains such as marketing, advertising, and scientific reports. The dataset comprises 199 fully annotated scanned forms divided into training/testing sets with the ratio 149/50. The configuration of the dataset enables the precise identification of key-value associations within forms, with elements such as questions, headers and other designated entities serving as keys, and their corresponding answers and related data serving as values. As such, the FUNSD dataset is an essential resource for studies focused on extracting relationships between entities within documents.

\section{Experiments}

\subsection{Experimental setup}

The architecture is described in detail in Section \ref{architecture}. We initialize our model using the LayoutLMv3 large model, which contains 357 million parameters and pre-trained on a subset of 11 million documents from IIT-CDIP Test Collection 1.0, representing 42 million scanned pages.
Each input is composed of one page with the associated text, image and spatial features. During training, those inputs are grouped into batches of size 2.\\

Text is tokenized with the same tokenizer as LayoutLMv3. Spatial coordinates are normalized to be between 0 and 1000 before being embedded. Images of the pages are resized to a squared format of 224x224 before DiT tokenization.\\

We apply our method on FUNSD and CORD datasets and evaluate them on the standard associated testing sets. To ensure a fair comparison, we use the same testing pipeline for both our method and GeoLayoutLM, and use the same datasets splits.

\subsection{Restriction on the Selection of Fathers (RSF): } 
\label{subsubsec:RSF}

For datasets where relations have a hierarchical structure, we select a set of predicted parent entities for each child entity. The post-processing method, Restriction on the Selection of Fathers (RSF), as proposed by GeolayoutLM \cite{geolayoutlm}, can be used to refine predicted relations and improve model performance. This method is particularly useful during inference when some entities are associated with multiple parent entities. Specifically, the \textit{j-th} entity is considered as a parent entity of the \textit{i-th} entity if and only if \( P_{i \rightarrow j} > 0.5\) and \(P_{i \rightarrow j} \) is  the maximum probability with a predefined margin \(\tau\): 
$$R_{ij} = 1\left( P_{i \rightarrow j} > 0.5\right) \times 1\left(\max_k  P_{i \rightarrow k} <  P_{i \rightarrow j} + \tau\right)$$
\\
\par To further enhance the RSF method, we use an additional variance loss also defined by GeolayoutLM \cite{geolayoutlm}. The underlying idea is that the probabilities of key entities should be closely aligned, implying that their variance should be minimized. During the fine-tuning, this variance loss is applied to those entities that are linked to more than one key entity.

\subsection{Experiments}

\begin{table}[t]
\centering
\setlength{\tabcolsep}{4pt} 
\renewcommand{\arraystretch}{1.2} 
\begin{tabular}{l|c|c|c}

\textbf{Method} & \textbf{\# Params} & \textbf{FUNSD} & \textbf{CORD} \\ \hline
BERT$_{LARGE}$ \cite{devlin2019bert} & 340M & 29.11 & - \\ 
LayoutLM$_{LARGE}$ \cite{Xu_2020} & 343M & 42.83 & - \\ 
StrucTexT \cite{li2021structext} & 107M & 42.83 & - \\ 
SERA \cite{zhang-etal-2021-entity} & - & 65.96 & - \\ 
LayoutLMv2$_{LARGE}$ \cite{xu2022layoutlmv2} & 426M & 70.57 & - \\ 
BROS$_{LARGE}$ \cite{hong2022bros} & 340M & 77.01 & - \\ 
GeoLayoutLM \cite{geolayoutlm} & 399M & 89.45 & 97.35 \\ \hline
LayoutLMv3$_{LARGE}$ & 357M & 82.38 & 96,92 \\ 
LayoutLMv3$_{LARGE}$ EEF + BBO & 357M & 88.16 & 97.14 \\ 
LayoutLMv3$_{LARGE}$ EEF + BBO + RSF & 357M & 89.64 & 96.91 \\ 
LayoutLMv3$_{LARGE}$ EM + BBO & 357M & 89.69 & \textbf{98.60} \\ 
LayoutLMv3$_{LARGE}$ EM + BBO + RSF & 357M & \textbf{90.81} & 98.48 \\ \hline
\end{tabular}
\vspace{4pt} 
\caption{Comparison of F1 scores for various models on the FUNSD and CORD datasets. Models are evaluated under different configurations to highlight their performance in document understanding tasks. \textbf{EEF}: Entity Extraction (EE) Joint Fine-tuning §\ref{subsubsec:E}, \textbf{EM}: Entity Marker §\ref{subsubsec:T}, \textbf{BBO}: Bounding Boxes Ordering §\ref{subsubsec:O}, \textbf{RSF}: Restriction on the Selection of Fathers §\ref{subsubsec:RSF}.}
\label{tab:f1_only_results}

\end{table}

In our study, we evaluated our model's performance against existing benchmarks on the CORD and FUNSD datasets, as shown in Table \ref{tab:f1_only_results}. To ensure a fair comparison, we re-evaluated GeoLayoutLM under its original setup on CORD dataset, since we used a different data modeling approach. The performance metrics for other models are as reported in their respective studies.\\

Our findings demonstrated that bounding boxes ordering significantly improves model performance, aligning with GeoLayoutLM's results. This underlines how the LayoutLMv3 model relies more on positional encoding than on layout information, and demonstrates the effectiveness of bounding box ordering in improving F1 scores to match GeoLayoutLM model that uses geometric pre-training. \\

Furthermore, incorporating entity semantic labels significantly improves performance on the CORD and FUNSD datasets, and outperforms joint learning on EE and RE. This result can be expected, as we provide the ground truth for labels that are directly used by the model. 

\subsection{Ablation Study}

To better understand the effectiveness of each strategy and their combination we perform multiple ablation studies.
\subsubsection{Effect of individual strategies:}

\begin{table}[t]
\centering
\setlength{\tabcolsep}{4pt} 
\renewcommand{\arraystretch}{1.2} 
\begin{tabular}{ l | l | c | c }
Model Ablation & F1 & Precision & Recall \\  \hline
Baseline & 82.38 & 80.54 & 84.31 \\ 
EE Joint Fine-tuning (EEF) & 83.56 {\color[HTML]{009901} (+1.18\%)} & 81.45 & 85.78 \\
Entity Marker (EM) & 86.48 {\color[HTML]{009901} (+4.10\%)} & 83.29 & 89.92 \\
Layout concatenation (LC) & 81.68 {\color[HTML]{FE0000} (-- 0.70\%)} & 79.44 & 84.04 \\
Bounding Boxes Ordering (BBO) & 86.91 {\color[HTML]{009901} (+4.53\%)} & 85.13 & 88.78 \\ 
Bounding Boxes Shuffling (BBS) & 85.22 {\color[HTML]{009901} (+2.84\%)} & 83.97 & 86.51 \\ \hline
\end{tabular}
\vspace{4pt} 
\caption{Separate impact of the five enhancement methods on relation extraction (RE) task for the \textbf{FUNSD} dataset compared to the baseline. Baseline is the LayoutLMv3 backbone with a bilinear prediction head.}
\label{tab:ablation study}
\end{table}
Our experimental results, summarized in Table \ref{tab:ablation study}, show the impact of individual contributions. Jointly fine-tuning the model on EE improves performance, but not as significantly as the entity marker strategy. Directly giving the model the entity types leads to a considerable improvement, especially in terms of recall, which is expected since knowing entity types aids in pre-identifying entities relationships.\\

Moreover, we can see that simply ordering the boxes along the y-axis gives the best results, especially in terms of precision, showing how much the model relies on positional encoding. By giving an order that better aligns with the reading order, we could considerably improve the model's performance.\\

Shuffling the bounding boxes during the training process within each batch seems to be a promising strategy. This implies that the order information could be dropped, and that the model would still learn to rely on the layout information rather than the order of the bounding boxes. While this strategy doesn't align well with the layoutLMv3 model's pre-training, which relies heavily on bounding box order, we suggest that this is what limits this shuffling approach, namely the gap between pre-training and tuning task. Thus, we would propose for future research a model that leverages the order of the tokens within a single segment, but uses no additional ordering information between different segments. \\

Finally, we observe that simply concatenating the layout information with the backbone embedding does not result in any improvement. We hypothesize that, because the information is redundant and already available in the token embeddings, introducing it before the relationship head increases model complexity and may lead to over-fitting. Another hypothesis is that the additional features did not go through non-linear operations and were not sufficiently enriched before being concatenated to the final token embedding.

\subsubsection{Effect of combining geometric and semantic strategies:}
\begin{table}[t]
\centering
\setlength{\tabcolsep}{4pt} 
\renewcommand{\arraystretch}{1.2} 
\begin{tabular}{ l | l | l | l }

  &  F1 &  Precision &  Recall \\ \hline
 EEF & 83.56 & 81.45 & 85.78 \\
 EEF + BBO &  88.16 {\color[HTML]{009901} (+4.60\%)} &  88.00 &  88.32 \\ 
 EEF + BBO + RSF &  89.64 {\color[HTML]{009901} (+6,08\%)} &  91.30 &  88.05 \\ 
 EEF + LC &  83.63 {\color[HTML]{009901} (+0.07\%)} &  81.59 &  85.78 \\ 
 EEF + LC + RSF & 86.11 {\color[HTML]{009901} (+2.55\%)} &  87.21 &  85.05 \\
 EEF + BBS &  85.06 {\color[HTML]{009901} (+1.50\%)} &  83.73 &  86.42 \\ 
 EEF + BBS + RSF &  87.93 {\color[HTML]{009901} (+4.37\%)} &  90.09 &  85.87 \\ 
 \hline 
 
 EM & 86.48 & 83.29 & 89.92 \\  
 EM + BBO &  89.69  {\color[HTML]{009901} (+3.21\%)}  &  87.91 &  91.54 \\ 
 EM + BBO + RSF &  90.81 {\color[HTML]{009901} (+4,33\%)} &  90.81 &  90.81 \\ 
 EM + LC &  86.76  {\color[HTML]{009901} (+0.28\%)} &  84.55 &  89.08 \\ 
 EM + LC + RSF &  90.03 {\color[HTML]{009901} (+3,55\%)} &  91.10 &  88.99 \\ 
 EM + BBS &  86.80 {\color[HTML]{009901} (+0,32\%)} &  84.62 &  89.08 \\ 
 EM + BBS + RSF &  89.13 {\color[HTML]{009901} (+2,65\%)} &  89.54 &  88.71 \\ \hline

\end{tabular}
\vspace{4pt} 
\caption{Impact of combining methods that add entity type information with geometric enhancements methods. \textbf{EEF}: Entity Extraction (EE) Joint Fine-tuning §\ref{subsubsec:E}, \textbf{EM}: Entity Marker §\ref{subsubsec:T}, \textbf{LC}: Layout Concatenation §\ref{subsubsec:L}, \textbf{BBO}: Bounding Boxes Ordering §\ref{subsubsec:O}, \textbf{BBS}: Bounding Boxes Shuffling §\ref{subsubsec:S} \textbf{RSF}: Restriction on the Selection of Fathers §\ref{subsubsec:RSF}.}
\label{tab:ablation study 2}
\end{table}

The second ablation study in Table \ref{tab:ablation study 2} showcases the synergistic effects of combining entity type information methods (EEF and EM) with geometric enhancements  methods (BBO, RSF, LC and BBS) on a model's performance in RE tasks. The key findings highlight that specific combinations, particularly those involving BBO and RSF, yield significant improvements in model performance. For instance, combining EEF or EM with BBO and RSF leads to the most notable enhancements, demonstrating the effectiveness of aligning data presentation with natural reading patterns and focusing on more probable relationship links.\\

The minimal impact observed from combining EEF or EM with LC suggests that simply adding layout information without strategic integration offers limited benefits. However, the incorporation of RSF across different combinations consistently results in performance gains, underlining RSF's critical role in enhancing the model's focus on relevant entity relationships. This indicates that while geometric enhancements and entity type information individually contribute to performance improvements, their strategic combination, especially with RSF, can significantly enhance the model's ability to extract and understand relationships between entities.

\section{Conclusion}
In this study, we explored the enhancement of Relation Extraction (RE) for Visually-Rich Documents (VRD) using a model based on LayoutLMv3, achieving or surpassing state-of-the-art results on FUNSD and CORD datasets without additional pre-training and with fewer parameters.\\

Through an extensive ablation study, this work highlights the significant impact of augmented features and modeling choices on the performance of RE models. The main findings indicate that the ordering of bounding boxes and the inclusion of entity semantic labels are pivotal factors in improving the RE task, underlining the importance of spatial and semantic information in understanding complex relationships.\\

Future research in this area could explore improved methods for encoding positional layout information, especially when concatenated with final token embeddings. Additionally, and more importantly, we believe there is a promising avenue for refining models for VRD to induce order only within text segments, and not between segments. This is because the usage or introduction of any explicit order could induce biases in the model, making it difficult to fine-tune for different or complex tasks.

%
%
%
%

\bibliographystyle{splncs04}
\bibliography{references}

\end{document}